\documentclass[letterpaper, 10 pt, conference]{ieeeconf}  % Comment this line out if you need a4paper

\IEEEoverridecommandlockouts                              % This command is only needed if 
\usepackage[pdftex]{graphicx}
\usepackage{xcolor}
\usepackage{soul}
\graphicspath{{../pdf/}{../jpeg/}}
\DeclareGraphicsExtensions{.pdf,.jpeg,.png}

\newcommand{\hcor}[1]{\textcolor[rgb]{0,0,0}{#1}}

\title{\LARGE \bf
Transparency evaluation for the
 Kinematic Design of the Harnesses through Human-Exoskeleton Interaction Modeling} %in LLE}

\author{Riccardo Bezzini$^{1}$, Carlo Alberto Avizzano $^{1}$, Francesco Porcini $^{1}$, and Alessandro Filippeschi$^{1}$% <-this % stops a space
\thanks{*This work was not supported by any organization}% <-this % stops a space
\thanks{$^{1}$ All authors are with the Institute of Mechanical Intelligence, and with the Department of Excellence in Robotics and AI, Scuola Superiore Sant’Anna, 56127 Pisa, Italy.}
}

\begin{document}

\maketitle
\thispagestyle{empty}
\pagestyle{empty}

%%%%%%%%%%%%%%%%%%%%%%%%%%%%%%%%%%%%%%%%%%%%%%%%%%%%%%%%%%%%%%%%%%%%%%%%%%%%%%%%
\begin{abstract}
Lower Limb Exoskeletons (LLEs) are wearable
robotic systems that provide mechanical power to the user.
Human-exoskeleton (HE) connections must guarantee the subsistence of the user’s natural behavior during the interaction, avoiding 
the exertion of 
undesired forces, i.e., the robot must be transparent.
Since transparency is an essential feature of exoskeletons' design, numerous works focus on its maximization, e.g., employing passive joints at the HE interfaces. 
Given the inherent complications of repeatedly prototyping and experimentally testing a device, modeling 
the exoskeleton and its physical interaction with the user emerges as an extremely valuable approach for assessing the design effects.
%the complication related to directly measuring 
% ((Despite transparency being one of the main goals of LLE design, only a few works have tackled the general problem of its maximization, and, to the author’s knowledge, none in dynamic settings.))
\hcor{This paper proposes a novel method to compare different exoskeleton configurations with a flexible simulation tool.}
% In this paper, a flexible simulation tool for comparing different exoskeleton configurations is implemented. The proposed method
\hcor{This approach }contemplates simulating the dynamics of the device, including its interaction with the wearer, to \hcor{evaluate} multiple connection mechanism designs \hcor{along with the kinematics and actuation of the LLE. This evaluation is based on the minimization of the interaction wrenches through an optimization process that includes the impedance parameters at the interfaces as optimization variables and the similarity of the LLE's joint variables trajectories with the motion of the %corresponding 
wearer's articulations.}
% and evaluate the resulting impact on the physical Human-Robot Interaction (pHRI). An optimization is performed to minimize the interaction wrenches, managing the impedance parameters at the interfaces. 
Exploratory tests are conducted using the Wearable Walker LLE \hcor{in different configurations} and measuring the interaction forces. % and measuring the interaction forces through load cells. %collected 
Experimental data are then compared to the optimization outcomes, proving that the proposed method provides contact wrench %or force
estimations consistent with the collected measurements and previous outcomes from the literature. 

%Moreover, the conclusions
%drawn at the design level are in accordance with the literature.

% \begin{itemize}
    % \item - Context: LLEs
    % \item - pHRI, transparency, and exo design
    % \item - Proposed solution and its relevance
    % \item - Brief: validity of results
% \end{itemize}
\end{abstract}

%IROS_EXO5 fa DESIGN OPTIMIZATION

\section{COPYRIGHT}
This paper has been accepted in the 2024 IEEE/RSJ International Conference on Intelligent Robots and Systems.

Copyright 2024 IEEE.  Personal use of this material is permitted.  Permission from IEEE must be obtained for all other uses, in any current or future media, including reprinting/republishing this material for advertising or promotional purposes, creating new collective works, for resale or redistribution to servers or lists, or reuse of any copyrighted component of this work in other works.

%%%%%%%%%%%%%%%%%%%%%%%%%%%%%%%%%%%%%%%%%%%%%%%%%%%%%%%%%%%%%%%%%%%%%%%%%%%%%%%%
\section{INTRODUCTION}
% \begin{itemize}
    % \item - Context: overview of LLEs and different applications
Lower limb exoskeletons are human-machine wearable systems that enhance the user’s performance, follow the human movement, and provide
 assistance. % to the wearer. 
 These devices usually consist of a main kinematic chain, whose
 joints aim to mimic human articulations, that runs parallel to the human limbs and is connected to it through the harnesses.
 Despite exoskeletons being an emerging and promising technology exploited for numerous purposes, spanning from human capabilities augmentation to rehabilitation
 or assistance, there
 are still significant limitations to deal with.
    % \item - Focus on pHRI, undesired int. forces, and Exo design
 Through the interaction %with the user, 
 resulting from the physical human-exoskeleton (HE) coupling, exoskeletons should follow the user’s motion
 without exerting unpredictable forces that may cause discomfort, pain, and damage. Therefore, while developing wearable exoskeletons, transparency should be the primary goal \cite{wang2023effect}. 
An ideally transparent exoskeleton
implements a perfect interaction with the user, who does
not feel the device as a hindrance but, on the contrary, 
cannot perceive its presence, even if it is providing the
demanded assistance. However, the employment of inappropriate HE connections inevitably leads to undesired interaction forces induced by the hyperstaticity deriving from the
human-exoskeleton kinematic mismatch. Minimizing undesired forces exerted on the human body is a fundamental concern in the physical Human-Robot Interaction (pHRI) analysis. Indeed, in the literature,
multiple strategies are suggested to deal with the interaction
forces at the HE interfaces, focusing on proper control strategies \cite{baris2024haptic, andrade2023experimental} and specific. %\st{definite} design choices \cite{wang2023analysis, cevik2021custom}. %, 8429068}. 
 However, to the author’s knowledge, there are no systematic approaches in dynamic settings for achieving transparency.
 
 % \item - BRIEF Exo design solutions for interaction forces: (general explanation) common solutions (NO HARNESS) with PROBLEMS
 %Many design solutions can be exploited for the minimization of the interaction wrenches. 
 Concerning mechanical design, several approaches can be exploited \cite{naf2018misalignment}.
 A conceivable procedure consists of introducing 
self-aligning mechanisms in series with the exoskeleton
joints. This strategy aims to delete the kinematic
mismatch’s effects by including additional joints or resorting to complex mechanical solutions that increase kinematic
compatibility \cite{kim2021bioinspired, hong2023design}. This method’s drawback is the resulting high complexity regarding actuation
 and control.
 Another widely used approach for achieving transparency involves adjustable length links and cuffs with modifiable positions, aiming at building a device suitable for anthropometrically different wearers \cite{zhang2023design}. Indeed, an exoskeleton should be designed to be compact, lightweight, and adjustable for a wide body-height range of users. Nevertheless, despite its unquestionable value, this solution alone does not ensure transparency.
 
 % OUT \item - Introduce JARRASSE MOREL solution (but static)
 % Among the numerous works from the literature that deal with the problem of transparency, while most of them focus on a specific device, in \cite{jarrasse2011connecting} the authors tackle the issue of undesired interaction forces from a general
 % perspective, even if their formulation builds on the assumption of static settings.
 Most of the works from the literature that deal with the problem of transparency focus on a specific device, limiting the generalizability of their conclusions. Conversely, this issue has been faced from a general perspective
 in \cite{jarrasse2011connecting}, even if their formulation builds on the assumption of static settings.
 The global goal is to design a mechanism at the human-robot interface such that all
 the forces generated on the human limb are controllable and there is no motion allowed to the
 exoskeleton while the human is still. The mentioned study demonstrates how to achieve a transparent design of the exoskeleton 
through passive compensation joints at the interfaces, as confirmed in the literature through the years \cite{lee2020design, sarkisian2021self}. 
The authors in \cite{7281266} and \cite{li2020influence}, for example, explore multiple design solutions for the harnesses collecting the interaction forces during various experiments. 
 Unfortunately, a significant limit of adding passive degrees of freedom (DoF) at the HE interface
 resides in the risk of harming the capability of the device to follow the wearer’s movement, as well as the inevitable increase in mass and mechanical complexity. Furthermore, despite the importance of experimental assessments for the evaluation of the HE interaction dynamics, prototyping and testing several designs require numerous and expensive experimental procedures.
 
It is worth noting that the reported approaches focus on the exoskeletons' mechanical design, usually not considering that the HE interaction forces are strongly related to the properties of the tissues involved in the contact at the interfaces.
%Certainly, practical tests are valuable for assessing the exoskeletons' performances, but 
Given the considerable variability of LLE designs and HE connection properties, modeling the coupled human-robot dynamics is becoming a widespread approach for evaluating an exoskeleton and deepening the knowledge of pHRI \cite{scherb2023modelling}.
Among the existing approaches for modeling HE interface properties, virtual springs (or spring-damper) are widely used for identifying the interaction dynamics \cite{guitteny2022dynamic}. However, many works only consider planar models in their analysis \cite{8744250, yan2023modelling}, while others limit the proposed investigation to static settings \cite{10304754}. Moreover, just a few implement a 3D (linear and rotational) contact model that includes the mechanisms of the harnesses, as well as the impedances at the interaction points,
 and, to the author's knowledge, none of them explore to this extent the HE interfaces in dynamic settings. 

In this paper, a three-dimensional articulated model of an LLE is formulated, including the harnesses, to simulate the dynamics of an exoskeleton during level walking. 
\hcor{In the proposed method, each harness terminates with a point that is connected to a point on the correspondent human bone of the virtual wearer through an impedance that includes the series of the human's soft tissues impedance and the impedance of the harness' interface with the human. This impedance is modeled through six spring-damper elements that allow the estimation of the contact wrenches. The method includes one harness for each of the femurs, tibias, and feet, and a seventh connection between the pelvic link of the LLE and the virtual wearer's back.}
% Each interaction point (the end of the kinematic chain representing the harnesses, not directly the exoskeleton link) is connected to the virtual wearer through six spring-damper elements (three linear and three rotational) for estimating the wrenches at the interfaces. 
% Moreover, a 3D impedance connects the pelvic link to the user. 
Each harness is composed of three prismatic and three revolute joints so that, by blocking some of the DoFs, multiple configurations of the connective mechanisms can be explored \hcor{using} the same model. 
\hcor{The motion of the virtual wearer is used to drive the simulation, whereas the estimated contact wrenches are used to compute a transparency metric to be optimized by varying the harness-human impedance parameters. This optimization allows a fair comparison of different HE harness designs. In fact, each design is considered with its best impedance parameters.}

In summary, this work proposes the formulation and preliminary validation of a method
 for evaluating the kinematic design of the HE interfaces in LLEs. The objective is to understand how many and which passive DoFs should be implemented in the connection mechanisms to improve transparency.
Once the dynamics equations of the HE interaction are formulated, a simulation is set up in Matlab-Simulink and used to solve an optimization problem whose
 variables are the impedance parameters that model the contact at the interfaces.
The main contributions of this paper are \hcor{hence}:
\begin{itemize}
    \item[-] \hcor{A flexible model} of the HE contact through 3D spring-damper elements, including both the harnesses mechanisms and the viscoelasticity of the involved tissues in the interaction dynamics.
    \item[-] The simulation of the HE interaction dynamics employing an LLE's 3D model with adjustable kinematic chains for exploring multiple harnesses configurations. %at thigh and shank.
    \item[-] The evaluation of different harnesses designs through optimization processes to maximize transparency.
    %explore the interaction dynamics and 
    \item[-] The preliminary validation of the proposed approach through comparisons with experimental data.
    %by comparing the experimentally measured interaction forces with those obtained from the proposed approach.
\end{itemize}
% IROS_EXO5 la fa a lista, ma include anche la validazione sperimentale...

% \begin{itemize}
%     \item - 3D dynamics simulation of the exo with adjustable structure of the harness
%     \item - 3D impedance at the interfaces
%     \item - experimental evaluation
% \end{itemize}

% \\
 The rest of the paper is organized as follows. In \hcor{section}
 \textit{Proposed Approach} the modeling procedure is explained, presenting the dynamics equations. Then the presented method for assessing the LLE design is reported. In section  \textit{Preliminary validation} the hardware employed and details about the experiments are presented.
 In \hcor{section \textit{Results and Discussion}} experimental and optimization data are compared and discussed, corroborating the validity of this work. Finally, \textit{Conclusions} terminates the paper, analyzing its main limitations
 and suggesting some plausible future directions.

\begin{figure}[thpb]
    \centering
    \includegraphics[scale=0.42]{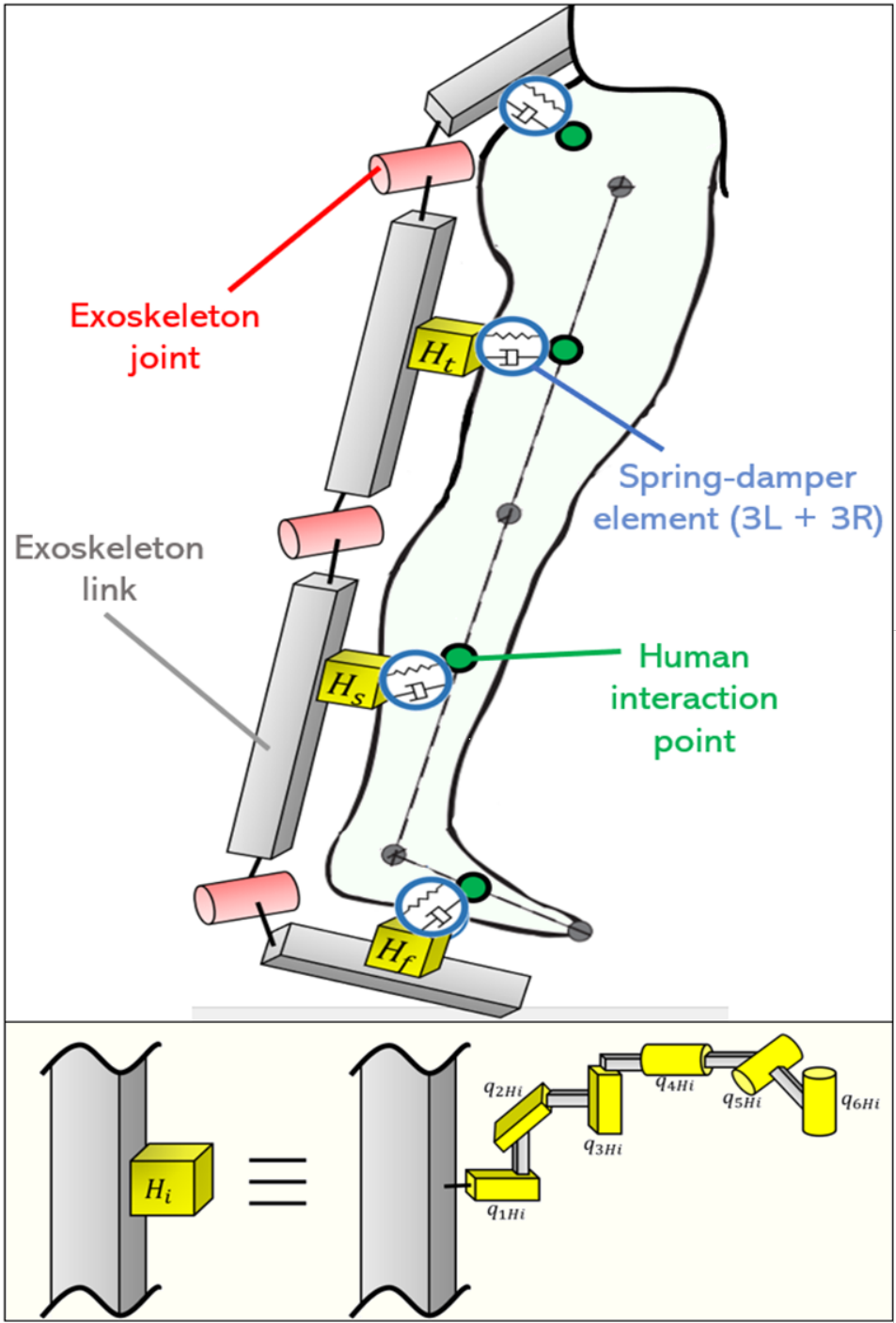}
    \caption{Sketched representation of the considered HE system. In the upper picture, one of the exoskeleton legs is depicted, including both the device's joints and links, together with the wearer's leg and the relative connection with the device. At each interface (besides the pelvic one) the harness mechanism links the LLE to the human through a three-dimensional virtual impedance. In the lower figure, the simplified representation of the harness is clarified, showing how each of these mechanisms includes six DoFs.}
    \label{F2}
\end{figure}

\section{PROPOSED APPROACH}
% \begin{figure*}[t]
%     \centering
%     \includegraphics[scale=0.4]{images/images/EXO_HUM_IROS6.pdf}
%     \caption{ciao}
%     \label{F1}
% \end{figure*}
% \begin{figure}[thpb]
%     \centering
%     \includegraphics[scale=0.37]{images/images/EXO_HUM_IROS7.pdf}
%     \caption{ciao}
%     \label{F2}
% \end{figure}

% \begin{itemize}
%     \item - general explanation: model (MATLAB-SIMULINK), exo+human, interaction model, optimization approach, objective
% LOOK METHODS AND ALGORITHM NELLA TESI
%     \item - model dynamics - equations (ALL terms)
%     \item - complete method scheme
%     \item - Optimization details (+schema)
% \end{itemize}
Achieving transparency is set as an optimization problem \hcor{that aims} to minimize \hcor{a metric based on} the HE interaction wrenches.
\hcor{These wrenches are computed by simulating the LLE dynamics, which includes the harnesses and the impedance at each contact point with the human. 
The impedances at the connection points include three rotational and three translational mutually orthogonal spring-damper elements. The optimization variables are the parameters that define the impedance of these human-robot connections at the interfaces.
Each harness is modeled as a six DoFs mechanism composed of three prismatic and three rotational joints as detailed in the following. Each harness joint includes a lumped spring-damper element whose parameters can be set to zero to set the DoF free or to a high value to lock the DoF.   
Therefore, the obtained simulation tool is flexible as it allows implementing different configurations for each harness simply by blocking some of its six possible DoFs. The motion of the LLE is driven by an imposed trajectory of the human bones. The relative motion of the human bones with regard to the harness at the contact points determines the contact wrenches. The optimization problem is completed by a set of constraints that guarantee that the motion of the LLE is coherent with the wearer's, avoiding eventual efficiency losses in transmitting assistive power to the user.}

\hcor{This optimization procedure allows the comparison of different harness designs. The optimization guarantees the fairness of the comparison, as each harness design is evaluated in the best configuration of HE impedance parameters, i.e., each configuration is evaluated assuming the best possible design for the interface of the harness with the wearer.}

\subsection{Human-exoskeleton coupled model}
\hcor{The human-robot coupled system (see Figure \ref{F2}) is modeled by rigid links, depicted as grey boxes, and joints, represented by parallelepipeds if prismatic or cylinders if revolute joints.}
% A representation of the human-robot coupled system is reported in Figure \ref{F2}, using links (depicted as gray
% parallelepipeds) and joints (sketched with parallelepipeds if prismatic and cylinders for
% revolving). 
The LLE is an articulated system with 42 DoFs: it includes 3 DoFs for each leg, i.e., flexion-extension for the hip and knee and plantarflexion-dorsiflexion at the ankle, and 6 DoFs for each harnesses located at thighs, shanks, and feet. The device is linked to the human limbs through the harnesses, modeled as six-joint mechanisms (three prismatic and three revolute joints), and virtual impedances at the interaction points, each composed of six spring-damper elements, three linear and three rotational. Each impedance represents
the series of the human’s soft tissues and the
impedance of the harness’ interface.
 An additional virtual impedance is employed to represent the HE interaction at the pelvis.
% \st{the viscoelastic contact properties at the interfaces.} 
% In figure \ref{F1} a sketched representation of the coupled system is reported, also depicting the kinematic composition of the mechanisms of the harnesses.
% The exoskeleton is considered linked to the human limbs through the harnesses, modeled as six joints serial
% kinematic chains, and virtual impedances at the interaction points (each composed of six mono-dimensional spring-damper elements, three linear and three rotational), used to represent the viscoelastic properties of the contact at the interfaces. 
\subsection{Dynamics equations}
The exoskeleton is considered grounded by its right foot. The objective is to build a simulation of a step during a gait cycle, and the simulated phenomenon corresponds to the phase during which the right foot is in contact with the ground and the left foot completes a swing movement. 
% Each human leg is
% considered composed of 8 serial revolute joints: 3 at the ankle, 2 at the knee, and 3 at the
% hip. The distances are considered null between the joints at the same human articulation. The
% human joint angles were extracted from [18].
% Even if this thesis focuses on the transparency during the gait cycle, simply changing the
% kinematic reference with human variables that describe another task the same approach could
% be exploited for drawing conclusions about the kinematic configuration of the exoskeleton and
% the harnesses.
% The interaction wrenches $w_{h_{i}}$ are computed based on the exoskeleton dynamics, considering that its motion depends on the interaction with the moving human legs, besides gravity and inertial terms. 
Said \textit{n} the number of joints and \textit{m} the number of connections, the exoskeleton dynamics is formulated as follows: %\begin{equation}\label{eq:TOTeq}
%     \mathbf{B\ddot{q}} + \mathbf{C\dot{q}} + \mathbf{g} = \tau + \Sigma_{h_{i} = 1}^{m} \mathbf{J^{T}_{h_{i}}w_{h_{i}}} - (\mathbf{K_{lock}q} + \mathbf{D_{lock}\dot{q}})  
%     %\hspace{20mm}where
% \end{equation}
\begin{equation}\label{eq:TOTeq}
    B\ddot{q} + C\dot{q} + g = \tau + \Sigma_{h_{i} = 1}^{m} J^{T}_{h_{i}}w_{h_{i}} - M_{lock}(q, q_0, \dot{q})%(K_{lock}q + D_{lock}\dot{q})  
    %\hspace{20mm}where
\end{equation}

where
% \begin{itemize}
%     \item$ \mathbf{q, \dot{q}, \ddot{q}} $ are the joints coordinates, velocities and accelerations respectively
%     \item $ \mathbf{B} = \mathbf{B(q)}$ is the inertia matrix of the exoskeleton 
%     \item $ \mathbf{C\dot{q}} = \mathbf{C(q, \dot{q})\dot{q}}$ is the vector of centripetal and Coriolis terms of the exoskeleton 
%     \item $ \mathbf{g} = \mathbf{g(q)}$ represents the vector of the gravitational contributions
%     \item $ \tau$ represents the vector of torques exerted by the exoskeleton's actuated joints
%     \item $\Sigma_{h_{i} = 1}^{m}\mathbf{J^{T}_{h_{i}}w_{h_{i}}}$ is the summation of the effects of the wrenches acting on the exoskeleton at the six interaction points 
%     \item $\mathbf{K_{lock}q} + \mathbf{D_{lock}\dot{q}} $ is a term that is used to assign the stiffness to certain DoFs of the harnesses, allowing to switch easily the configuration of the harnesses in the simulation
% \end{itemize}
\begin{itemize}
    \item[-] $q, \dot{q}, \ddot{q} \in R^{42\times1}$ are the joints coordinates, velocities and accelerations respectively
    \item[-] $B =B(q)$ is the inertia matrix of the exoskeleton 
    \item[-] $C\dot{q} =C(q, \dot{q})\dot{q}$ is the vector of centripetal and Coriolis terms of the exoskeleton 
    \item[-] $g =g(q)$ is the vector of the gravitational contributions
    % \item[-] $ \tau$ is the vector of torques exerted by the exoskeleton's actuated joints
    \item[-] $ \tau$ is the actuated joints' torques vector % vector of torques exerted by the exoskeleton's actuated joints
    \item[-] $\Sigma_{h_{i} = 1}^{m}J^{T}_{h_{i}}w_{h_{i}}$ is the sum of the effects of the wrenches acting on the LLE at the interaction points 
    % \item[-] $ K_{lock}q +D_{lock}\dot{q} $ is a term used to assign the stiffness to certain DoFs of the harnesses, allowing to switch the configuration of the harnesses in the simulation
    \item[-] $ M_{lock}(q, q_0, \dot{q})$ is a term used to assign the stiffness to certain DoFs of the harnesses, allowing to switch the configuration of the harnesses in the simulation
\end{itemize}

% la metterei prima perchè così motivo il calcolo dei wrenches dopo

 % spiegazione percentuali riferite a \cite{pons2008wearable}

% Even if this thesis focuses on the transparency during the gait cycle, simply changing the
% kinematic reference with human variables that describe another task the same approach could
% be exploited for drawing conclusions about the kinematic configuration of the exoskeleton and
% the harnesses.

% % MDPI paper
% the motion-tracking system that is worn by the user. In the current implementation,
%  the Xsens MVN-Link-Biomech tracking system (Enschede, 7521, The Netherlands).

\subsubsection{Actuated joints torques}
\hcor{Actuated joint torques, i.e., the term $\tau$ of equation \ref{eq:TOTeq}, allow for the implementation of the LLE control strategy in the simulation. In the formulation adopted in this paper the friction and ripple effects are assumed to be compensated at the lower control level, whereas $\tau$ may represent several LLE control strategies.}

% \begin{figure*}[thpb]
%     \centering
%     \includegraphics[scale=0.38]{images/images/schema_fin_largeSL.png}
%     \caption{Block diagram of the proposed approach. The \textit{OFFLINE} section regards the acquisition of the joint variables during a gait cycle through the XSENS suit and the exoskeleton initialization phase. The \textit{ONLINE} part depicts the data flow in the Simulink system during the optimization process.}
%     \label{scheme}
% \end{figure*}

\begin{figure*}[thpb]
    \centering
    \includegraphics[scale=0.31]{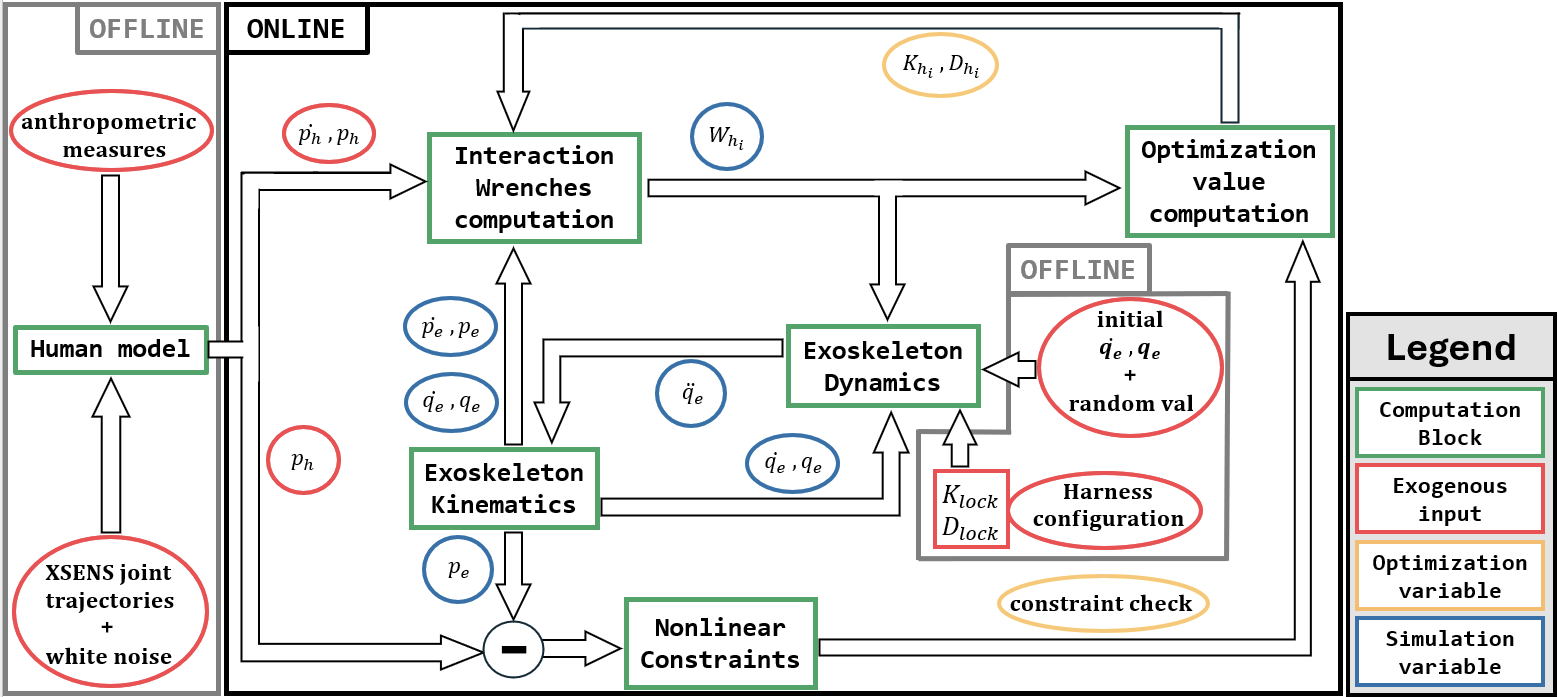}
    \caption{Block diagram of the proposed approach. The \textit{OFFLINE} section regards the acquisition of the joint variables during a gait cycle through the XSENS suit and the exoskeleton initialization phase. The \textit{ONLINE} part depicts the data flow in the Simulink system during the optimization process.}
    \label{scheme}
\end{figure*}

\subsubsection{Human-robot interaction}

The pHRI contribution is embodied by the term $\Sigma_{h_{i} = 1}^{m}J^{T}_{h_{i}}w_{h_{i}}$ of equation \ref{eq:TOTeq}. Each wrench is computed according to the poses and twists at the two ends (human and robot) of the spring-damper elements:%\ref{eq:Wint}:
% \begin{equation} \label{eq:Wint}
%         \mathbf{w_{h_{i}}} =  \mathbf{K_{h_{i}}(p_{exo}^{h_{i}} - p_{human}^{h_{i}}}) + \mathbf{D_{h_{i}}(\dot{p}_{exo}^{h_{i}} - \dot{p}_{human}^{h_{i}}})
% \end{equation}
% \begin{itemize}
%     \item $ \mathbf{K_{h_{i}}} \in \mathbf{R}^{6\times6}$ is the diagonal matrix that defines the stiffness (translational and rotational) of the virtual springs at the end of the \textit{i-th} harness
%     \item $ \mathbf{D_{h_{i}}} \in \mathbf{R}^{6\times6} $ is the diagonal matrix that defines the damping coefficient (transnational and rotational) of the virtual dampers at the end of the \textit{i-th} harness
%     \item $\mathbf{ p_{exo}^{h_{i}} - p_{human}^{h_{i}} }$ is the difference between the poses (position and orientation) of human and robot interfaces points coupled by the \textit{i-th} harness
%     \item $ \mathbf{\dot{p}_{exo}^{h_{i}} - \dot{p}_{human}^{h_{i}} }$ is the difference between the twists of human and robot interfaces points coupled by the \textit{i-th} harness.
% \end{itemize}
\begin{equation} \label{eq:Wint}
        w_{h_{i}} =  K_{h_{i}}(p_{h}^{h_{i}} - p_{e}^{h_{i}}) + D_{h_{i}}(\dot{p}_{h}^{h_{i}} - \dot{p}_{e}^{h_{i}})
\end{equation}

where
\begin{itemize}
    \item[-] $ K_{h_{i}}$ and $ D_{h_{i}} \in R^{6\times6}$ are the diagonal matrices that define the translational and rotational stiffness and damping coefficients of the virtual spring-damper elements at the interface of the \textit{i-th} harness
    % \item[-]  \in R^{6\times6} $ is the diagonal matrix that defines the damping coefficient (translational and rotational) of the virtual dampers at the end of the \textit{i-th} harness
    \item[-] $ p_{h}^{h_{i}}$, $p_{e}^{h_{i}}, \dot{p}_{h}^{h_{i}}$ and $\dot{p}_{e}^{h_{i}} \in R^{6\times1}$ are respectively the poses ($p$, position and orientation) and twists ($\dot{p}$) of the human ($h$) and exoskeleton's ($e$) interaction points coupled by the \textit{i-th} harness $h_i$ (with $h_0$ identifying the pelvis).
    % \item[-] $ \dot{p}_{h}^{h_{i}}$ and $\dot{p}_{e}^{h_{i}} \in R^{6\times1}$ are the twists of human and robot interfaces points coupled by the \textit{i-th} harness
\end{itemize}

\subsubsection{Kinematic reference}

%The objective is to build the simulation of a step during level walking. 
To compute the human interaction points poses $p_h^{h_i}$ and twists $\dot{p}_h^{h_i}$ the Xsens MVN-Link-Biomech tracking system (Enschede, 7521, The Netherlands) is used. The proprietary software (XSens MVN Analyze) extracts lower limb joints trajectories which are then exported in Matlab. The gait cycle portion we are interested in includes the $[12\div50] \%$ fraction for the stance leg and the $[62\div100] \%$ for the swing leg (the percentages are referred to \cite{pons2008wearable}).
 The variables from the considered time interval are given as input to a kinematic model of the human lower limbs that includes three perpendicular revolute DoFs at each of the hip, knee, and ankle articulations (18 DoFs). The anthropometric measures can be changed in the model, hence, the same joint variables can be used to run simulations that investigate how the exoskeleton impacts different wearers.

\subsubsection{Harness configuration}

To block the selected harnesses' DoFs, whose structure comprises three prismatic and three revolute joints, in equation \ref{eq:TOTeq} the term 
\begin{equation}
    M_{lock} = K_{lock}(q - q_0) + D_{lock}\dot{q}
\end{equation}
is used. It allows imposing a certain stiffness on the joints that must be locked. $K_{lock}$ and $ D_{lock} \in R^{n\times n} $ are diagonal matrices, representing stiffness and damping coefficients. By modifying their values the harness's kinematic design can be changed, e.g., setting the $K_{lock}$ and $D_{lock}$ diagonal elements to zero leaves the corresponding harness' DoFs free.
% leaving free the DoFs corresponding to the matrices' null diagonal elements. 
The harness configurations are identified by a vector \textbf{[\textit{xyz}]} that illustrates the number of free DoFs implemented in the connection mechanism at the thigh $(\textbf{\textit{x}})$, shank $(\textbf{\textit{y}})$, and foot $(\textbf{\textit{z}})$ interfaces, with $(\textbf{\textit{x}}, \textbf{\textit{y}}, \textbf{\textit{z}}) \in [0\div 6]$.
For example, the arrangement \textbf{[0 1 0]} represents a configuration with zero DoFs implemented in the thigh and foot's harnesses, and a single DoF connection at the shank.

\subsection{Optimization}

The simulation loop system, based on the equation \ref{eq:TOTeq} and symbolized in figure \ref{scheme}, was implemented in Matlab-Simulink and then used for the optimization process. The \textit{GlobalSearch} algorithm \cite{ugray2007scatter} (Global Optimization Toolbox) was employed.
 Since the goal is to evaluate the kinematic structure of the harnesses reducing the interaction wrenches, the cost function to be minimized is computed as:
\begin{equation}\label{eq:OPTeq}
    \lambda = w^{T}_{list}W_{opt}w_{list}.
\end{equation}
The diagonal matrix $W_{opt} \in R^{(m\times6)\times(m\times6)}$ contains the weights for the {\it m} interaction wrenches $w_{h_{i}}$, that are incorporated in $w_{list} \in R^{(m\times6)\times(1)}$.
% where 
% \begin{itemize}
%     \item[-] $W_{opt} \in R^{(m\times6)\times(m\times6)}$ is a diagonal matrix with the weights for the interaction wrenches
%     \item[-] $w_{list} \in R^{(m\times6)\times(1)}$ contains the {\it m} wrenches $w_{h_{i}}$
% \end{itemize}  
 The algorithm runs the simulation loop minimizing the cost function calculated in equation \ref{eq:OPTeq}, \hcor{using the elements of $K_{h_{i}}$ and $D_{h_{i}}$ (see equation \ref{eq:Wint}) as optimization variables.} The impedance at the pelvis is fixed and therefore not included in the optimization process.

The optimization aims at reducing the interaction wrenches, and since the stiffness and damping parameters are the optimization variables, the obvious solution would be a null impedance at the interfaces. This would correspond to null interaction forces, that is the meaningless case of an exoskeleton not connected to the human limbs. 
However, transparency should be achieved without hindering the assistive effectiveness of the device, therefore the \textit{nonlinear contraints} were exploited to ensure the HE connection.

% \begin{figure*}[thpb]
%     \centering
%     \includegraphics[scale=0.265]{images/images/schema_fin_large.png}
%     \caption{Block diagram of the proposed approach. The \textit{OFFLINE} section regards the acquisition of the joint variables during a gait cycle through the XSENS suit and the exoskeleton initialization phase. The \textit{ONLINE} part depicts the data flow in the Simulink system during the optimization process.}
%     \label{scheme}
% \end{figure*}

\subsubsection{Nonlinear constraints for tracking guarantee}
The objective function is minimized such that
$c(x) \leq 0$, where %$x$ is the optimization variable (the vector containing all the stiffness and damping parameters of the impedances at the interfaces) and  
$c(x)$ is computed as a function of the distances between the exoskeleton and the human interaction points:% during the simulation:
\begin{equation}
    c(x) =  \Sigma^{N}_{i = 1}(  \Sigma^{k}_{j = 1}DISTbool(j))
\end{equation}
where $N$ represents the number of time steps in the simulation, while $k$ stands for the number of elements of the vector \textit{DISTbool} that contains all the distances between the exoskeleton and human's interaction points at each instant $i$. 
$DISTbool$ elements are computed at each instant $i$ as:
\begin{equation} \label{eq:F_rep}
DISTbool(j) = \left\{\begin{array}{ll}
1 & if \mid D_{sim}(j) - D_{th}(j) \mid \geq  0 \\
0 & if \mid D_{sim}(j) - D_{th}(j) \mid <  0
\end{array}\right.
\end{equation}

The vector $D_{sim} \in R^{18\times1}$ contains the six three-dimensional distances between the exoskeleton and the human interaction points (thighs, shanks, and feet), while $D_{th}\in R^{18\times1}$ is the vector composed of the maximum values that the distances are allowed to reach during the gait cycle.
%If the threshold is exceeded, the corresponding element of $DISTbool$ is different from zero, increasing the value of $c$ making the simulated solution unsuitable.

% \subsubsection{Implementation}

% The simulation loop system, based on \ref{TOTeq}, was implemented in Matlab-Simulink and then used
% for the optimization process. In particular, The GlobalSearch algorithm  \cite{ugray2007scatter} (Global Optimization Toolbox) was used.

% and its 
% function, that 
% uses a scatter-search mechanism for generating start points, analyzes them, and rejects those that are unlikely to improve the best local minimum.
% A scheme that visually renders the data flow of the proposed method is presented in figure \ref{scheme}

% \section{EXPERIMENTAL SETUP}
\section{PRELIMINARY VALIDATION}

% spiegazione dei risultati presentati, quali sono, perchè etc

% Experimental tests and simulations have been conducted. Some of the results obtained from the
% optimization (led with a human model that respects the anthropometric measures of the tested
% subject) have been preliminarily investigated through experimental validation, while others
% are considered for a comparison with outcomes from the literature. The selected device is the Wearable Walker, a
% powered LLE able to both assist and augment human capabilities \cite{camardella2021gait}. It is composed of 7 links
% connected by 2 active DoFs per leg (hip and knee flexion/extension) and 1 non-actuated DoFs
% per leg (ankle flexion/extension). The exoskeleton is equipped with load cells on the thigh and
% tibia human-robot interfaces to measure the forces shown in Figure 6. These forces have
% been collected by testing one subject wearing the exoskeleton and walking on a treadmill.

% Experimental and simulated tests have been conducted. Some of the results obtained from the
% optimizations (led with a human model that respects the anthropometric measures of the tested
% subject) have been preliminarily investigated through experimental validation, while others
% are compared with results from the literature. 

The proposed approach was tested firstly by comparing the optimizations' output with existing findings in the literature. Then, two available LLE configurations were adopted to compare the simulations' outcomes with experimental data gathered on the real device, finally exploring an additional configuration to be possibly implemented on the exoskeleton.
% Here, the employed hardware is presented, together with the procedures for experimental data acquisition and some details about the optimization settings.

% \subsection{Experimental setup}
   \begin{figure*}[thpb]
      \centering
      \includegraphics[scale=0.36]{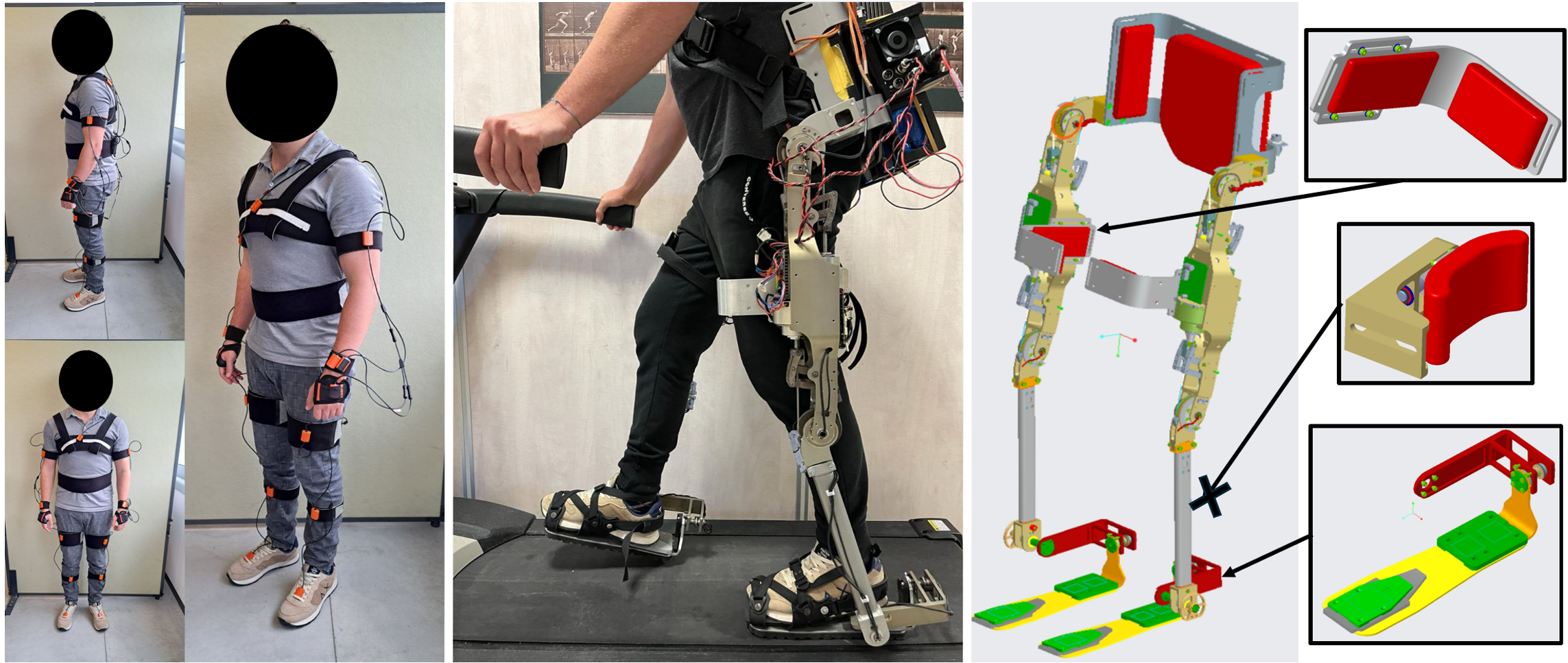}
      \caption{Starting from the left, three photos of the user wearing the Xsens suit are presented. In the central figure, the user is walking on a treadmill with the Wearable Walker. On the right figure a CAD representation of the \textbf{[261]} device is reported, also displaying the harnesses' models.}
      \label{P_C_X}
   \end{figure*}

\subsection{Lower Limb Exoskeleton System}
The Wearable Walker is a powered LLE able to assist and augment human capabilities \cite{camardella2021gait}. It weighs 19 kg and is composed of 7 links
connected by 2 active DoFs (hip and knee flexion/extension) and 1 non-actuated DoF
per leg (ankle dorsi/plantarflexion). It is equipped with two load cells on each thigh and tibia interface to evaluate both the interaction forces on the sagittal plane and perpendicular to it.  Through the active joints, friction and ripple are compensated by the low-level control, while the high-level controller compensates for the gravitational and inertial terms.
%in two directions at each harness interface. One ($y$) is perpendicular to the sagittal plane, while the other direction ($z$) lays on the sagittal plane and is perpendicular to the exoskeleton's link.

\subsubsection{Harnesses Configurations}
The available harnesses on the Wearable Walker can be arranged in two layouts that were tested experimentally. Moreover, based on the LLE kinematics and verified outcomes from the literature, a third arrangement, i.e. \textbf{[3 3 2]}, was selected to explore a solution to be possibly implemented on the LLE. 
\begin{itemize}
    \item[-] \textbf{[0 1 0]}: the thigh and foot harnesses can be considered as embedments, while the shank interface leaves free the rotation of the limb around his longitudinal axis.
    \item[-] \textbf{[2 6 1]}: the shank is not connected to the exoskeleton, while the thigh harness leaves free the translation along the longitudinal axis and the rotation around the axis perpendicular to the sagittal plane. The foot harness leaves free the inversion-eversion.
    \item[-] \textbf{[3 3 2]}: the thigh and shank’s rotation and translation along the limbic axis are freed. The rotation is implemented in this virtual solution because of the absence of internal rotation in the device's hip and knee articulations, while the translation is left free so that the user can easily adjust the position of the harness along the limb. The
third conceded DoF is the rotation around a horizontal axis parallel to the sagittal plane, aiming to supply for the absence of the adduction/abduction movement.
Excluding the translation, the same DoFs are implemented in the foot harness.
\end{itemize} 
\subsection{Method Exploration}
\begin{figure}[thpb]
      \centering
      \includegraphics[scale=0.22]{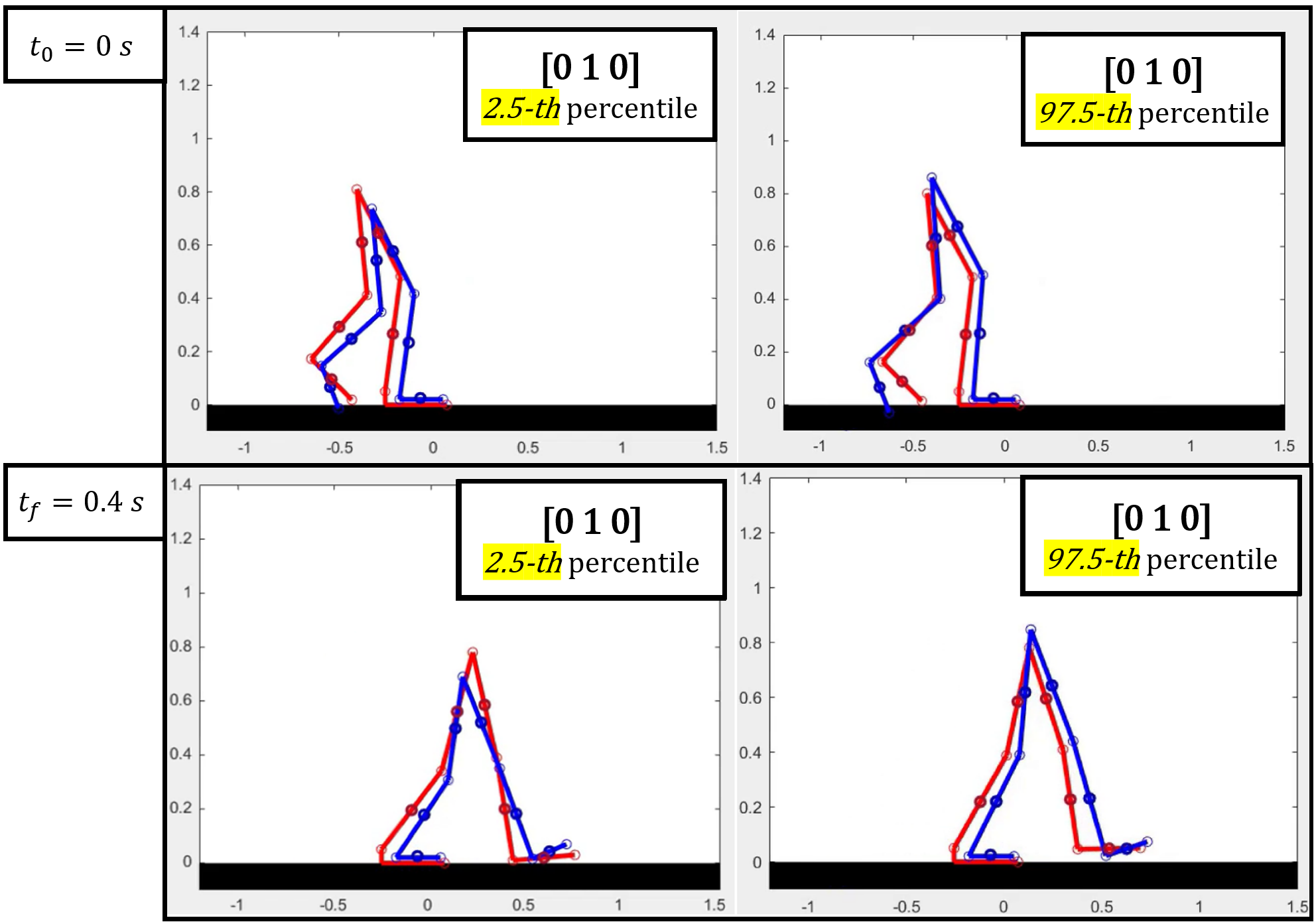}
      \caption{Stick-plots of the starting ($t_0$) and final ($t_f$) instants of simulations that employ the \textit{2.5-th} and \textit{97.5-th} percentiles anthropometric measures for the human model. 
      % This sketch on the sagittal plane was built on Matlab just to provide a straightforward display of the simulated phenomenon.
      }
      \label{perc}
   \end{figure}

% - da RESULTS: prendere la parte introduttiva in cui si spiega l'esplorazione (con dettagli numerici)
% - Explain differences between optimizations
Before comparing experimental and simulated data, an exploration of the method is conducted to prove its robustness by introducing three sources of variability: 
\begin{itemize}
    \item[-] The variation of the anthropometric measures of the virtual wearer (belonging to the $2.5$-$th$, $50$-$th$ and $97.5$-$th$ percentiles, see Figure \ref{perc}).
    \item[-] The addition of random variations (with maximum value $\gamma = 0.1$ $rad$) on the exoskeleton initial conditions, that is to say on the joint variables.
    \item[-] The introduction of Gaussian white noises on the human kinematic references through the Matlab function \textit{awgn()}, using a \textit{signal-to-noise} ratio equal to 30.
\end{itemize}

% Additive white Gaussian noise (AWGN) is a simple noise model that represents electron motion. The noise gets added to the signal and is called white because it is spectrally flat across the entire bandwidth. The noise is Gaussian because its amplitude can be modeled with a normal probability distribution.

% The AWGN channel is often used to model a satellite communications channel, since that channel typically does not suffer from common terrestrial impairments like fading, multipath, and interference. An AWGN channel serves as a good starting point for the analysis of terrestrial wireless links because it establishes a best-case bound on the bit error rate performance of a terrestrial link.
Using the \textbf{[0 1 0]} configuration, different combinations of variability sources were used to run the optimizations, determining both the interaction wrenches and the exoskeleton kinematic performances under these mutable conditions. %, for an assessment of the harness's role in the interaction with the wearer. 
Regarding the interaction wrenches, the mean values and variances are computed employing the RMSs of each component of the wrench for every optimization cycle. The results are compared with outcomes from the literature to evaluate their adequacy in terms of module and deviation. For the kinematics assessment, the effects of the nonlinear constraints limiting the maximum distances are tested by monitoring the differences between the exoskeleton and human joint motion during the simulation. The vector $D_{th}$ from equation \ref{eq:F_rep} is set with all the elements equal to $10$ $cm$.
% This analysis was conducted to certify the robustness of the proposed approach to variations in the simulation settings.
% Furthermore, the kinematics of the simulated exoskeleton
% is monitored for a complete assessment of the harness's role in the interaction with the wearer. As explained in the previous section, nonlinear constraints are used for limiting the maximum distances between the HE interaction points during the simulations. The vector $D_{th}$ has been chosen with all its elements equal to $10$ $cm$. During the analysis of the HE movement, the differences between human and exoskeleton joints motion throughout the simulation are reported to investigate the connections' role on the faculty of the exoskeleton to follow the user's movement.
% transparency level
% entailed by the analyzed LLE configurations and assessing the validity of the optimizations' results.
%Along with the considerations about pHRI, 
\subsection{Experimental assessment}
One healthy volunteer aged 31, tall 1.78 m, and with a mass of 76 kg participated in this preliminary experiment. The procedure includes a human joint variable acquisition without LLE to run the optimizations followed by a session for measuring the interaction forces with the LLE.
\subsubsection{Human variables acquisition}
The Xsens MVN suit (see Figure \ref{P_C_X}) is composed of IMUs (sampling frequency of $240$ $Hz$) placed on the human body segments through elastic bands. \hcor{The participant wore the suit and performed level walking. As mentioned above, only the samples of definite gait cycle's percentages were} employed to compute the twists and poses of the virtual wearer. 
\subsubsection{Interaction forces measurements}
The mentioned interaction forces have been collected by testing one participant wearing the Wearable Walker while walking on a treadmill. 
The \hcor{experiment was conducted for both the available harness configurations, and consisted of} 120-second long sessions of level
 walking. Overall, 4 acquisitions have been conducted, 2 for each of the explored design arrangements.
 Figure \ref{P_C_X} a shows the user walking on a treadmill while wearing the Wearable Walker during one of the performed acquisitions.

\subsubsection{Validation metrics}

A comparison between the RMS (root mean square) values of the experimental and simulated forces is used to evaluate the capacity of the proposed approach to determine the impact of a certain design on the pHRI. Once the validity of the proposed approach has been preliminarily confirmed by the comparison with experimental data, the \textbf{[3 3 2]} configuration was investigated. For this purpose, similarly to the procedure carried out in the method exploration, the interaction wrenches at the interfaces in the three analyzed cases are compared, as well as the differences between the human and exoskeleton joints motion.

\section{RESULTS AND DISCUSSION}

% Before comparing experimental and simulated data, an exploration of the method is conducted to prove its robustness by introducing three sources of variability: a
% variation of the anthropometric measures of the virtual wearer (belonging to the \textit{2.5-th}, \textit{50-th} and \textit{97.5-th} percentiles), random variations of the exoskeleton initial conditions, and the addition of white noises on the human kinematic references.
% Then the RMS (root mean square) values of the experimental and simulated forces are used to evaluate the capacity of this approach to assess the impact of a certain design on the pHRI. 
% % transparency level
% % entailed by the analyzed LLE configurations and assessing the validity of the optimizations' results.
% %Along with the considerations about pHRI, 
% Furthermore, the kinematics of the simulated exoskeleton
% is monitored for a complete assessment of the harness's role in the interaction with the wearer. Therefore, the differences between human and exoskeleton joints motion during the simulation are reported to investigate the connections' role on the faculty of the exoskeleton to follow the user's movement.

\begin{figure}[thpb]
    \centering
    \includegraphics[scale=0.365]{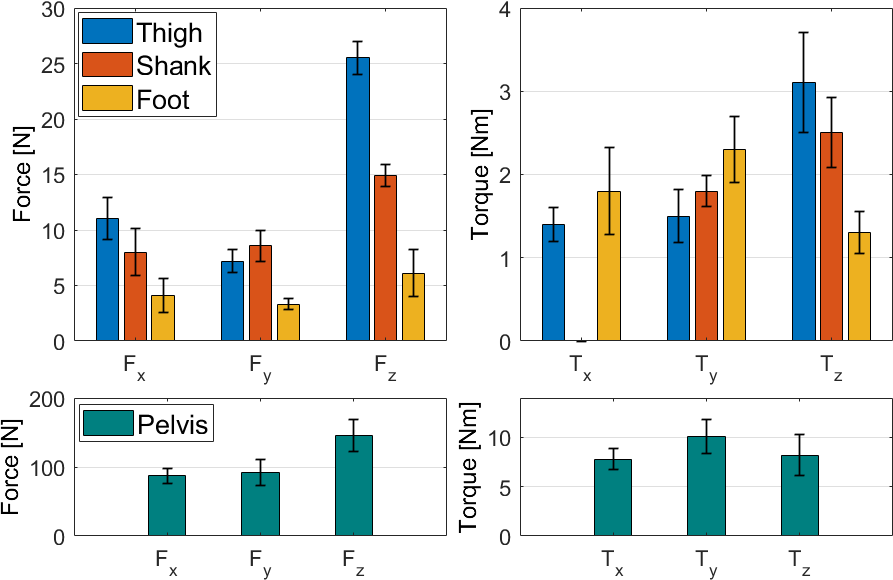}
    \caption{The simulated interaction wrenches in the \textbf{[0 1 0]} case. The mean absolute values and standard deviations are computed based on the RMSs for each component of the wrenches from optimization processes that differ from each other for the inclusion of all the mentioned variability sources.}
    \label{EXPL1}
\end{figure}

% \subsection{Method exploration}
 
\begin{figure}[thpb]
    \centering
    \includegraphics[scale=0.22]{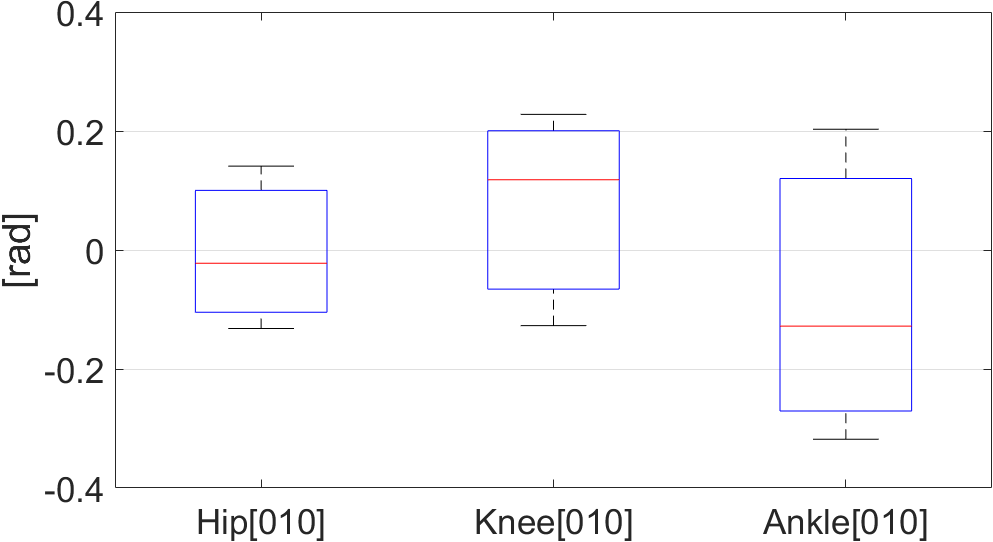}
    \caption{Boxplots of the differences between the angular position of the exoskeleton joints with respect to the virtual human joints motion (case \textbf{[0 1 0]}). The employed optimization processes are distinguished from each other through the inclusion of all the mentioned variability sources.}
    \label{BP010}
\end{figure}

The interaction wrenches from the optimizations are reported in figure \ref{EXPL1} to present the effects of the possible combinations of the mentioned sources of variability in the \textbf{[0 1 0]} configuration, employed as an example to test the robustness of the proposed approach to variations on the inputs.
Moreover, Figure \ref{BP010} reports the boxplots of the differences between the human and exoskeleton joint variables to assess whether the simulated LLE follows the wearer's motion. 
% configurations of the exoskeleton are explored, comparing the values obtained from a "standard" optimization (conducted using exoskeleton nominal initial conditions and anthropometric
% measures from the tested subject) with the forces calculated as the mean of data from six different
% simulations (combining human percentile and noise alterations). On the horizontal axis the
% force’s subscript denotes if the force is measured at the thigh (T) or shank (S), and what axis
% of the simulation’s base frame is parallel to the considered force.
% \begin{figure}[thpb]
%     \centering
%     \includegraphics[scale=0.2]{images/images/schema_1801+sat.png}
%     \caption{METHOD EXPLORATION: Here the bar graphs of the simulated interaction wrenches are reported, regarding the case \textbf{[0 1 0]}. The mean values and standard deviations are computed based on the RMS values for each component of the wrench, computed through the results of multiple optimization processes. Each of these optimization cycles differ from each other for the white noise additions on the initialization and reference values and the variability of the simulated anthropometric measures.}
%     \label{EXPL1}
% \end{figure}
Both the interaction wrenches and the HE relative motion are comparable with the literature. Indeed, in \cite{yan2023modelling} the authors explore the HE interaction forces and the absolute tracking errors, deriving values that are consistent with those reported in this paper. In \cite{nair2020performance}, the reported angular displacements are comparable with the ones presented in this work. Furthermore, interaction forces and torques collected during multiple gait cycles with the iT-Knee in \cite{saccares2016knee} are similar to the ones obtained in this work. 

%Therefore it can be stated that t
% Thus this approach has proven
% quite robust to plausible changes in the simulated conditions.
%Moreover, from a preliminar comparison with the literature, the numerical values are coherent with the mentioned works from the literature that report interaction wrenches in LLEs. \cite{yan2023modelling} \cite{sarkisian2021self}
% \subsection{Harness evaluation}

\begin{figure}[thpb]
    \centering
    \includegraphics[scale=0.28]{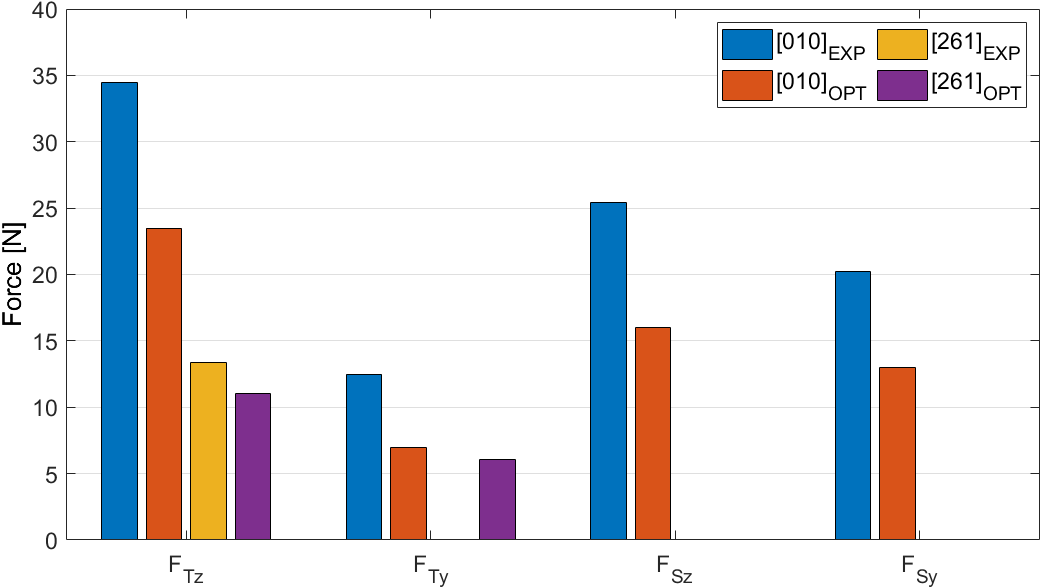}
    \caption{The thigh and shank interaction forces' RMSs from the experimental acquisitions (\textit{EXP}) and the forces from the optimizations (\textit{OPT}) with the actual user's anthropometric measures and white noise added to the human reference motion and random variations of the LLE's initial conditions.}
    \label{MvsO}
\end{figure}

Once verified the robustness of the approach, the measured forces at thighs (T) and shanks (S) are compared with the optimizations results (see Figure \ref{MvsO}).
 The missing forces correspond to the freed DoFs in the harnesses (as aforementioned, friction is not considered in this model).
It can be noted that the experimental forces are quite accurately emulated in the simulations. Indeed, the proportions between the measured forces can be extracted from the simulation outcomes \hcor{for the different harness configurations}. Given the obvious differences between the simulation and the experiment (e.g. the wearer's motion), the optimization does not aim at an accurate replication of the experimental results, but at the evaluation of different configurations and harness designs. Results reported in Figure \ref{MvsO} show the potentiality of the approach as a tool to evaluate a specific harness mechanism by minimizing the interaction impedances. Since the actual harnesses were not designed based on these optimizations, the simulated forces have overall lower values.

\begin{figure}[thpb]
    \centering
    \includegraphics[scale=0.365]{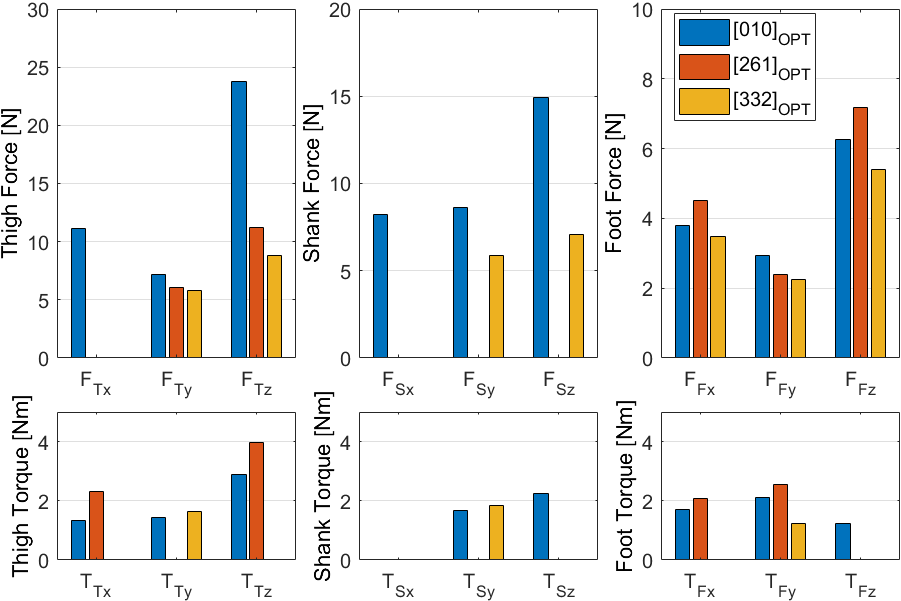}
    \caption{Interaction forces RMSs from the optimizations regarding three harnesses distributions, including all the variability sources.}
    \label{T_S_TOT}
\end{figure}

\begin{figure}[thpb]
    \centering
    \includegraphics[scale=0.21]{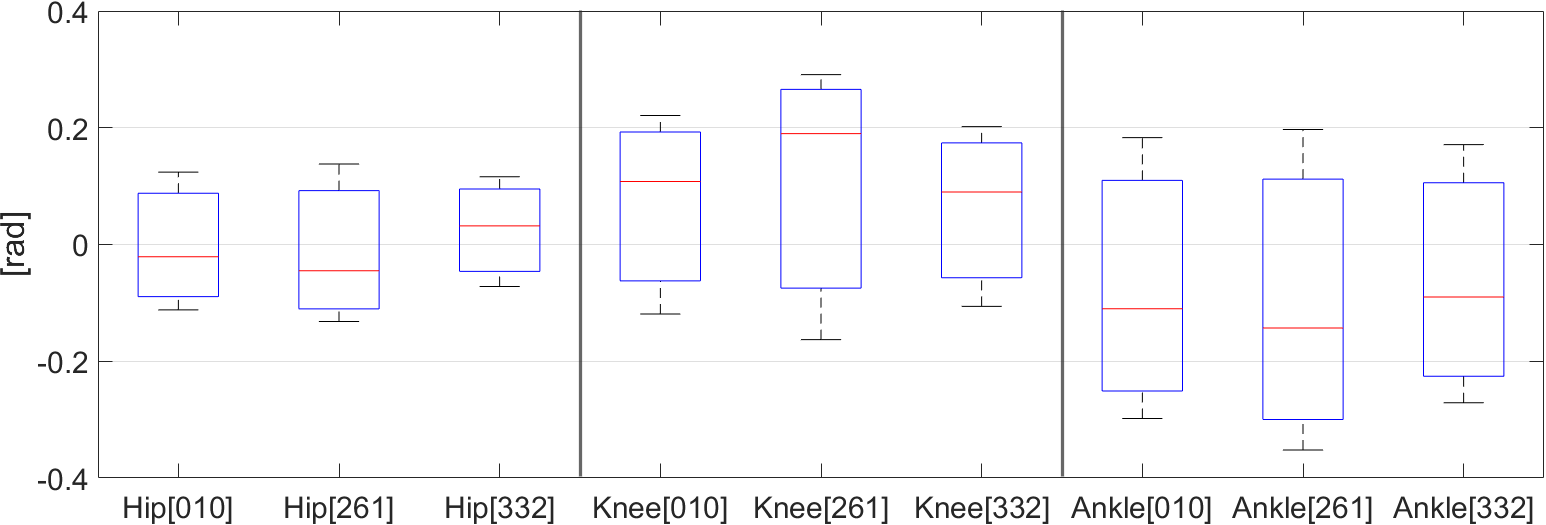}
    \caption{The boxplots of the joint motion differences between LLE and virtual wearer in the simulated cases including all the variability sources.}
    \label{BPtot}
\end{figure}

The proposed approach was finally employed to evaluate a possible improvement in the current connection mechanisms through the \textbf{[3 3 2]} arrangement. %, that has been tested with respect to the two existing configurations. 
In figure \ref{T_S_TOT}, the simulated interaction wrenches resulting from different harnesses are reported, while in figure \ref{BPtot} the human-exoskeleton relative motion is analyzed.
Introducing passive DoFs at the interfaces improves transparency \cite{jarrasse2011connecting}, indeed the higher wrenches are the ones in \textbf{[0 1 0]}. The absence of wrenches at the shanks in \textbf{[2 6 1]} might be a desirable feature, but from the kinematic performances, it can be noted that the \textbf{[2 6 1]} arrangement confers a lower efficacy to closely follow the user, which may be undesirable. Moreover, the absence of interaction at the shanks induces higher wrenches at the foot in \textbf{[2 6 1]}. Regarding the other two configurations,
 their kinematic performances are quite similar, but \textbf{[3 3 2]} assures lower interaction wrenches, proving to be a valuable mechanical solution to improve transparency on the Wearable Walker.

To sum up, the method results to be robust to reasonable input variations, consistent with experimental data and findings from the literature, and reliable for assessing the influence of the harnesses' kinematic design on transparency.

\section{CONCLUSIONS}

%%%%%%%%%%%%%% future works
% In order to improve the accuracy of the results, some features of the implementation reported in
% the previous chapters can be changed. First of all, the accuracy of the model can be increased
% including the contribution of the motors in the dynamics. Once the exoskeleton kinematic
% configuration has been defined, simulators like AnyBody or OpenSim could be used for further
% confirmation of the results.
% The model of the human legs could also be refined, studying in deep the joints distribution and the geometry of the kinematic chain that describes the human limbs. By the same
% reasoning, other models for the impedance at the interfaces could be explored.
% Concerning the experimental validation, increasing the number of tested subjects would
% improve the foundation of the conducted examinations.
% Aiming at corroborating the proposed approach for more then one application, other tasks
% could be simulated and experimentally tested, and same goes for attempts with different exoskeletons.

This paper presented a novel approach for enhancing transparency in LLEs by the evaluation of different harness designs through a flexible simulation and optimization tool.
 The robustness of the method and its coherence with experimental data has been preliminarily assessed, showing the potentiality of the approach. 
The outcomes are consistent with the collected data, hence the investigations corroborate the worth of this method for evaluating the design of the connection mechanisms to maximize transparency, encouraging its use to improve pHRI by exploring novel harnesses design solutions. Further experimental assessments will be carried out to strengthen these results, and the simulation itself will be improved by including the models of the motor and allowing for more complex LLE kinematics.

\bibliographystyle{ieeetr}
\bibliography{mybib}

\end{document}